\documentclass[lettersize,journal]{IEEEtran}

\usepackage{amsmath,amsfonts}
\usepackage{pifont}
\usepackage{array}
\usepackage[caption=false,font=normalsize,labelfont=sf,textfont=sf]{subfig}
\usepackage{textcomp}
\usepackage{stfloats}
\usepackage{url}
\usepackage{verbatim}
\usepackage{graphicx}
\usepackage{cite}
\usepackage{epsfig}
\usepackage{amssymb}
\usepackage{booktabs}
\usepackage{enumitem}
\usepackage{multirow}
\usepackage{lipsum}
\usepackage{xcolor}
\usepackage[export]{adjustbox}
\definecolor{codegreen}{rgb}{0.0,0.6,0.0}
\definecolor{blue}{rgb}{0.21,0.49,0.74}

\usepackage[ruled,vlined,linesnumbered]{algorithm2e}
\makeatletter
\newcommand{\algorithmfootnote}[2][\footnotesize]{%
  \let\old@algocf@finish\@algocf@finish
  \def\@algocf@finish{\old@algocf@finish
    \leavevmode\rlap{\begin{minipage}{\linewidth}
    #1#2
    \end{minipage}}%
  }%
}
\makeatother
\makeatletter
\DeclareRobustCommand\onedot{\futurelet\@let@token\@onedot}
\def\@onedot{\ifx\@let@token.\else.\null\fi\xspace}

\def\etc{\emph{etc}\onedot}

\makeatother
\usepackage[pagebackref,breaklinks,colorlinks]{hyperref}

\usepackage[capitalize]{cleveref}
\crefname{section}{Sec.}{Secs.}
\Crefname{section}{Section}{Sections}
\Crefname{table}{Table}{Tables}
\crefname{table}{Tab.}{Tabs.}

\newcommand{\lzl}[1]{\textcolor{black}{#1}}

\begin{document}

\title{TrackSSM: A General Motion Predictor by State-Space Model.}

\author{Bin Hu,
        Run Luo,
        Zelin Liu,
        Cheng Wang,
        Wenyu Liu,
\IEEEcompsocitemizethanks{
\IEEEcompsocthanksitem B. Hu, R. Luo, Z. Liu, C. Wang, W. Liu are with Huazhong University of Science and Technology, China.\protect\\
\IEEEcompsocthanksitem Corresponding author: Wenyu Liu. Email: \url{liuwy@hust.edu.cn}.\protect\\
}
}

\markboth{Journal of \LaTeX\ Class Files,~Vol.~14, No.~8, August~2021}%
{Shell \MakeLowercase{\textit{et al.}}: A Sample Article Using IEEEtran.cls for IEEE Journals}

\IEEEpubid{0000--0000/00\$00.00~\copyright~2021 IEEE}



\IEEEtitleabstractindextext{
\begin{abstract}
Temporal motion modeling has always been a key component in multiple object tracking (MOT) which can ensure smooth trajectory movement and provide accurate positional information to enhance association precision. However, current motion models struggle to be both efficient and effective across different application scenarios. To this end, we propose TrackSSM inspired by the recently popular state space models (SSM), a unified encoder-decoder motion framework that uses data-dependent state space model to perform temporal motion of trajectories. Specifically, we propose Flow-SSM, a module that utilizes the position and motion information from historical trajectories to guide the temporal state transition of object bounding boxes. Based on Flow-SSM, we design a flow decoder. It is composed of a cascaded motion decoding module employing Flow-SSM, which can use the encoded flow information to complete the temporal position prediction of trajectories. Additionally, we propose a Step-by-Step Linear (S$^2$L) training strategy. By performing linear interpolation between the positions of the object in the previous frame and the current frame, we construct the pseudo labels of step-by-step linear training, ensuring that the trajectory flow information can better guide the object bounding box in completing temporal transitions. TrackSSM utilizes a simple Mamba-Block to build a motion encoder for historical trajectories, forming a temporal motion model with an encoder-decoder structure in conjunction with the flow decoder. TrackSSM is applicable to various tracking scenarios and achieves excellent tracking performance across multiple benchmarks, further extending the potential of SSM-like temporal motion models in multi-object tracking tasks. Code and models are publicly available at \url{https://github.com/Xavier-Lin/TrackSSM}.
\end{abstract}

\begin{IEEEkeywords}
2D multi-object tracking, state space model (SSM), temporal motion model, hidden state, flow information.
\end{IEEEkeywords}
}

\maketitle
\IEEEdisplaynontitleabstractindextext

\section{Introduction}\label{intro}
\IEEEPARstart{M}{odeling} complex the linear and nonlinear motion has always been a key issue in multi-object tracking (MOT) tasks~\cite{vandenhende2021multi}. For scenarios with intense motion, such as dance scenes~\cite{dancetrack}, sports~\cite{sportsmot}, and autonomous driving~\cite{bdd100k}, robust and efficient motion modeling has become an essential component of high-performance trackers. Although previous trackers \cite{fairmot,jde,bytetrack,motrv2,p3aformer,MeMOTR} have achieved advanced performance on multiple benchmarks, robust and efficient motion modeling for various different scenarios remains a significant challenge.

Successful motion modeling needs to ensure the following two points: 1) robustness to diverse motion patterns, 2) high inference efficiency. The current mainstream motion models in MOT adopts the Kalman filter~\cite{kf}, which is based on the assumption of constant velocity and is data-independent. It typically use a equation of constant velocity motion to compute the prior state of a trajectory and update this state with matched observations to predict the trajectory position at the next time step. However, when the actual movement of a target significantly deviates from the motion prior, it can lead to erroneous trajectory associations. Some approaches \cite{transtrack,transcenter,trackformer,motr} use attention-based autoregressive methods for temporal propagation of trajectories and demonstrate superior performance in nonlinear motion scenarios. With the increase in the number of tracking targets, attention-based autoregressive modeling leads to a quadratic growth in computational cost. Additionally, some methods \cite{centertrack,siammot} employ convolutional neural networks (CNN) for temporal autoregressive modeling, integrating them with detection networks within the same framework to form siamese or shared-parameter networks. Although such approaches improve computational efficiency, they can lead to feature conflicts between tracking and detection tasks, resulting in weaker detection performance.

Recently, state space models (SSM) \cite{s4,s5} have achieved widespread success in efficiently handling long sequence tasks, owing to their efficient computation of sequential information and effective state transition modeling. Inspired by SSM, we propose a unified motion framework based on data-dependent SSM, named TrackSSM. It follows an encoder-decoder architecture. The encoder is composed of stacked naive Mamba \cite{mamba} modules, which aggregate the position and motion representations of historical trajectories to obtain the trajectory flow information. The decoder consists of cascaded motion decoding modules from our proposed Flow-SSM, which can utilize the flow information obtained from the encoder to guide the temporal position prediction of the current frame trajectories. Additionally, to improve the accuracy of trajectory position prediction, we propose a Step-by-Step Linear(S$^2$L) training strategy. By linear interpolating the trajectory positions between the current frame and the previous frame, we construct step-by-step linear training pseudo labels, guiding the bounding box to complete temporal transitions in a progressive linear manner. Compared to Mamba, we parameterize the SSM using the flow information encoded from historical trajectories, resulting in Flow-SSM. It effectively handles various linear and nonlinear motion target position transitions. 
\IEEEpubidadjcol
Benefiting from the efficient computation of the Mamba module, the inference speed of TrackSSM with the YOLOX-l \cite{yolox} detector can reach up to 27.5 FPS, surpassing most attention-based temporal autoregressive motion models \cite{transtrack,trackformer,MeMOTR,motr,motrv2}.

With a fixed detector model and hyper-parameter configuration, the TrackSSM with the YOLOX-x \cite{yolox} detector achieves comparable performance to the baseline ByteTrack \cite{bytetrack}, which uses Kalman Filter(KF) \cite{kf} as the motion model, on the MOT17 \cite{mot16} test set. On the DanceTrack \cite{dancetrack} test set, ByteTrack integrated with TrackSSM achieves a tracking performance of 57.7 HOTA \cite{hota}, a gain of +10.9 HOTA compared to the baseline. On the SportsMOT~\cite{sportsmot} test set, ByteTrack with TrackSSM achieves a tracking performance of 74.4 HOTA, a gain of +11.0 HOTA compared to the baseline. Notably, TrackSSM paired with the detector can infer at real-time speed and incurs less computation overhead. Experimental results on different benchmarks demonstrate that TrackSSM has the potential to become a universal motion framework in multi-object tracking tasks.

Our contributions are summarized as follows:

\begin{itemize}
\item{We propose Flow-SSM, a module that guides the temporal state transition of object bounding boxes using flow information generated by the encoder.}

\item{
Based on Flow-SSM, we design the flow decoder, which can utilize the flow information from the trajectories of historical frames to perform the temporal position prediction.}

\item{
We propose a step-by-step linear(S$^2$L) training strategy. By performing linear interpolation on trajectory positions, we construct step-by-step linear training pseudo labels, ensuring that the flow information from historical frame trajectories can more accurately guide the object bounding box in performing temporal predictions.}

\item{
Combining the above designs, we propose TrackSSM, a simple and effective motion model with the encoder-decoder structure. TrackSSM is applicable to various tracking scenarios and achieves excellent performance across multiple tracking benchmarks.}
\end{itemize}

\section{Related work}\label{related}
\subsection{2D Multi-Object Tracking}
Current mainstream 2D multi-object tracking methods can be categorized into Tracking-By-Detection(TBD) paradigms and Joint Detection and Tracking(JDT) paradigms. Most TBD methods \cite{Bewley2016_sort, deepsort, motdt, jde, fairmot, bytetrack, BoT-SORT, strongsort, sparsetrack, mtracker} typically use Kalman Filter(KF) as the motion model, which predicts the prior position of the trajectory at the next time step and associates it with the corresponding detection. Although TBD methods have achieved impressive performance on multiple tracking benchmarks, their performance is somewhat limited by hyper-parameters and specific scenarios. To compensate for the shortcomings of the KF motion model in modeling complex scenarios, appearance features are introduced as an important association metric. These features help recall lost trajectories when occlusion and loss occur. While powerful appearance models \cite{fastreid, torchreid,zhou2019osnet,zhou2021osnet} are beneficial for accurate tracking, the efficiency of the tracker decreases as the number of objects in the scene increases. To improve the robustness of trackers across different scenarios and reduce the number of hyper-parameters, JDT methods have been proposed to simultaneously perform object detection and trajectory temporal position prediction tasks. CenterTrack \cite{centertrack} tracks objects as points, detecting and tracking the center points of objects while predicting their temporal displacements. SiamMOT \cite{siammot} uses a siamese network to jointly optimize the detection and tracking tasks within the same framework, performing temporal regression of trajectory positions via the tracking network. TransTrack \cite{transtrack} is the first to introduce the Transformer architecture into tracking algorithms, constructing a tracking method dependent on track queries. It represents trajectories as queries to predict the position of the previous frame trajectory in the current frame. TrackFormer \cite{trackformer} is the first to propose a tracking method based on continuous temporal autoregression of trajectory queries, allowing trajectory queries to possess true continuous time tracking capabilities. MOTR \cite{motr} represents both detection and tracking tasks as a set prediction problem within a single stage, achieving truly end-to-end multi-object tracking.

\subsection{Methods for Motion Modeling in MOT}
Motion modeling in multi-object tracking can be divided into two categories: heuristic motion models and learnable motion models. Heuristic methods typically use fixed motion priors and a set of hyper-parameters to control the trajectory motion process, with the Kalman Filter (KF) \cite{kf} being a typical example. While KF motion models have been successful in most tracking benchmarks \cite{mot15,mot16,mot20}, they can lead to failed tracking results in benchmarks with more intense motion. To address the limitations of the KF, GIAOTracker \cite{giaotracker} proposed the NSA Kalman filter motion model, which aims to adaptively adjust the noise scale (the covariance information of the object) based on the quality of object detection, achieving success in multi-object tracking benchmarks that involve complex motion. Other approaches \cite{BoT-SORT,sparsetrack} use camera motion compensation to mitigate the intensity of object motion, followed by naive Kalman filtering for motion prediction. Both naive and NSA Kalman filters come with a large number of hyper-parameters, which poses a potential risk of being limited to specific types of scenarios. As a result, learnable motion models have gradually attracted researchers' attention, thanks to their data-driven nature. Tracktor \cite{Tracktor} is the first tracker to propose a learnable motion model, using trajectory boxes as Regions of Interest (RoI) \cite{girshick14CVPR} in each frame, extracting corresponding RoI features to regress the trajectory boxes to the current frame. MotionTrack \cite{motionTrack} learns the representation of the trajectory at historical moments and uses it to predict the movement of the trajectory at the next moment. Trajectory autoregressive models based on self-attention mechanisms \cite{attention,vit} can partially overcome the challenge of motion modeling for occluded objects. However, they can cause conflicts between detection and tracking tasks during the tracking process, weakening detection performance. DiffMOT \cite{diffmot} constructs a temporal diffusion motion model to replace the KF, viewing the regression process of the trajectory box from the previous frame to the current frame as a diffusion-denoising \cite{diffusion} process, achieving some success across different tracking benchmarks. Despite these advances, efficient and robust general motion modeling remains an area that requires further exploration.

\subsection{The State Space Model}
State space models (SSM) \cite{s4,s5} generally refer to a class of models that utilize hidden states for sequential autoregression of objects. The current widely-used state space model is S4 \cite{s4}, whose autoregressive process can be described as follows:
\begin{equation}
\begin{aligned}
\label{s4}
h_t &= \mathbf{\overline{A}}h_{t-1} + \mathbf{\overline{B}}x_t, \\
y_t &= \mathbf{C} h_t + \mathbf{D} x_t,
\end{aligned}
\end{equation}
Where $x_t$ represents the input signal, $h_t$ is the hidden state, and $y_t$ is the output signal. $\overline{A}$, $\overline{B}$, $C$ and $D$ are four matrix parameters, with $\overline{A}$ and $\overline{B}$ being discrete parameters that can be obtained through zero-order hold(ZOH) and Euler discretization rules. Although S4 offers a concise and efficient solution for long-sequence modeling, further exploration is needed for modeling longer and more complex sequences. S5 \cite{s5} introduces efficient parallel scanning and MIMO-SSM into the S4 layer, further enhancing the sequence modeling capabilities of SSM. Recently, \cite{mamba} introduces a data-driven SSM layer, referred to as the Mamba module. The core design of the Mamba module involves parameterizing the $\Delta$, $B$ and $C$ matrix parameters using representations extracted from sequence data. Benefiting from its data-driven nature, the Mamba module is better equipped to capture sequence features and enhance long-sequence modeling capabilities. Notably, the Mamba module scales linearly with sequence length, while maintaining low memory overhead and high inference efficiency. In this work, we design a motion prediction model based on the SSM structure and thoroughly explore the potential of SSM architecture in temporal prediction tasks.

\section{method}\label{meth}
\begin{figure*}[!t]
\centering
\includegraphics[width=0.96\linewidth]{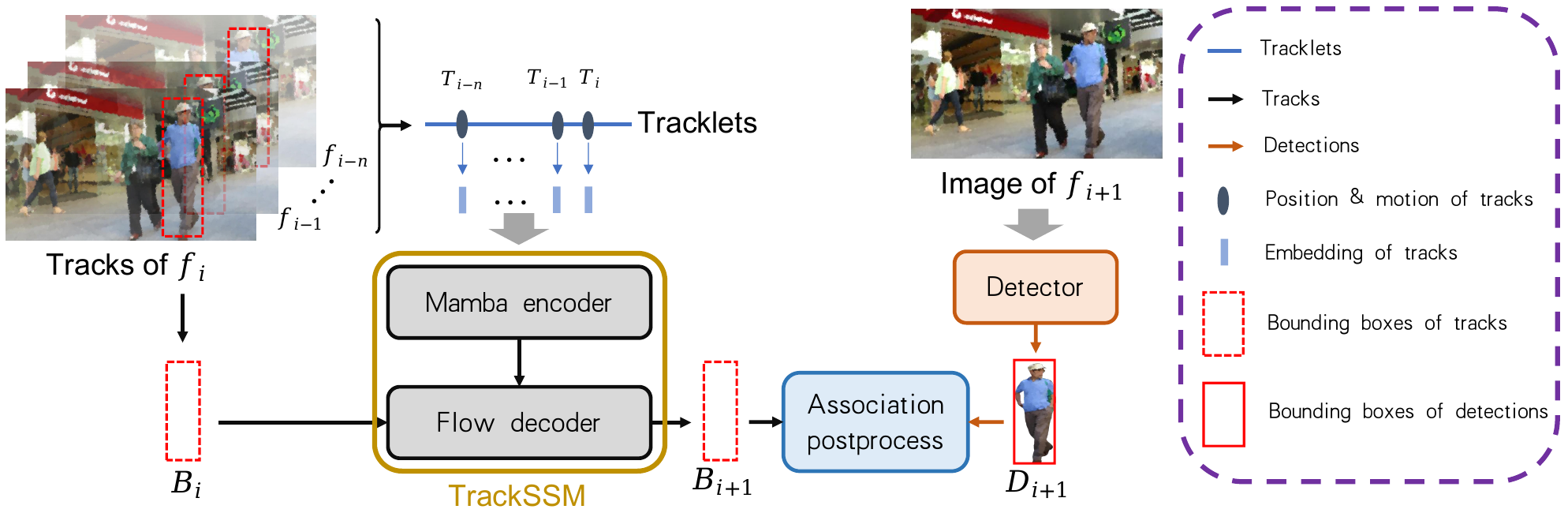}
\caption{
The overall tracking framework with the TrackSSM motion model, where TrackSSM consists of the Mamba encoder \cite{mamba} and Flow decoder, is capable of performing temporal predictions on trajectories. The legend information is located in the box on the right side.
}
\label{framework}
\end{figure*}
\subsection{General Framework}
The overall tracking framework via the TrackSSM motion model is shown in \cref{framework}. Given the position and motion information of a trajectory $\{T_{i-k}=(x_c,y_c,w,h,\Delta_{x}, \Delta_{y}, \Delta_w, \Delta_h)\}^{0}_{k=n}$ for $n$ historical frames, we encode the trajectory information $T_{i-k}$ at each time step into trajectory embeddings $\mathcal{T}_{i-k} \in \mathbf{R}^{m}$, forming a sequence of trajectory embeddings $\{\mathcal{T}_{i-k}\}^{0}_{k=n}$. The embedding sequence is then fed into a naive Mamba encoder \cite{mamba}, with the output representation at the final time step serving as the motion flow information of the trajectories, which we refer to as the flow feature $\mathcal{F} \in \mathbf{R}^{m}$. The flow feature contains abundant historical information about the trajectory with position and motion. Subsequently, we use the flow feature $\mathcal{F}$ as guidance, feeding it into a designed flow decoder to guide the trajectory box $B_i$ in predicting its position $B_{i+1}$, which can obtain a prediction track box at the time $(i+1)$. During the tracking phase, the trajectory prediction box $B_{i+1}$ is associated with the detection box $D_{i+1}$ obtained by the detector, and the association process is similar to that in ByteTrack \cite{bytetrack}.

\begin{algorithm}[h]
\SetAlgoLined
\DontPrintSemicolon
\SetNoFillComment
\footnotesize
\KwIn{
Track position embeddding $\mathcal{E}_i:(\texttt{B},\texttt{D})$;  
Flow features $\mathcal{F} : (\texttt{B},\texttt{M})$; Hidden state $h : (\texttt{B},\texttt{D},\texttt{N})$}

\KwOut{Track position embeddding $\mathcal{E}_{i+1} : (\texttt{B},\texttt{D})$; 
Hidden state $h' : (\texttt{B},\texttt{D},\texttt{N})$}
\tcc{Parameterize data-independent matrices}
$\mathbf{A}:(\texttt{D},\texttt{N}) \leftarrow $ Parameter\;
$\mathbf{D}:(\texttt{D},) \leftarrow $ Parameter\;
\tcc{Parameterize data-dependent matrices via flow features}
\textcolor{codegreen}{$\mathbf{\Delta}:(\texttt{B},\texttt{D}),\mathbf{B}:(\texttt{B},\texttt{N}),\mathbf{C}:(\texttt{B},\texttt{N}) \leftarrow \mathbf{Linear}(\mathcal{F})$\;}

\tcc{Discretize}
$\mathbf{\overline{A}}:(\texttt{B},\texttt{D},\texttt{N}) \leftarrow \mathbf{Exp}(\mathbf{\Delta} \bigotimes \mathbf{A})$\;
$\mathbf{\overline{B}}:(\texttt{B},\texttt{D},\texttt{N}) \leftarrow \mathbf{\Delta} \bigotimes \mathbf{B}$\;

\tcc{Running SSM}
\textcolor{codegreen}{$\mathcal{E}_{i+1}:(\texttt{B},\texttt{D}), h':(\texttt{B},\texttt{D},\texttt{N}) \leftarrow \mathbf{SSM}(\mathbf{\overline{A}},\mathbf{\overline{B}}, \mathbf{C},\mathbf{D})(h, \mathcal{E}_i)$\;}
 
Return: $\mathcal{E}_{i+1}, h'$
\caption{Flow-SSM}
\algorithmfootnote{The key steps of Flow-SSM are in \textcolor{codegreen}{green}.}
\label{flow-ssm}
\end{algorithm}
\subsection{Flow-SSM}
To achieve the process of using flow features to guide trajectory boxes for temporal prediction, we design Flow-SSM. The algorithm pseudo-code is shown in \cref{flow-ssm}, where $\texttt{B}$ represents the batch size, $\texttt{D}$ denotes the dimension of the state space model, and $\texttt{N}$ is the state dimension. We adopt the zero-order hold(ZOH) method to discretize the parameters $\mathbf{A}$, $\mathbf{B}$ into $\mathbf{\overline{A}}$, $\mathbf{\overline{B}}$ through the time scale $\mathbf{\Delta}$, following the method \cite{mamba}. Compared to the Mamba module, Flow-SSM parameterizes the $\mathbf{\Delta}$, $\mathbf{B}$ and $\mathbf{C}$ matrices using flow features $\mathcal{F}$ to achieve temporal guidance of the trajectory embeddings $\mathcal{E}_i$. Unlike traditional state space models \cite{s4,s5,mamba}, Flow-SSM operates on a sequence length of only 1. It means that Flow-SSM does not perform sequence modeling like traditional state space models but instead predicts the state of the input signal at the next time step.
\begin{figure*}[!ht]
\centering
\includegraphics[width=0.96\linewidth]{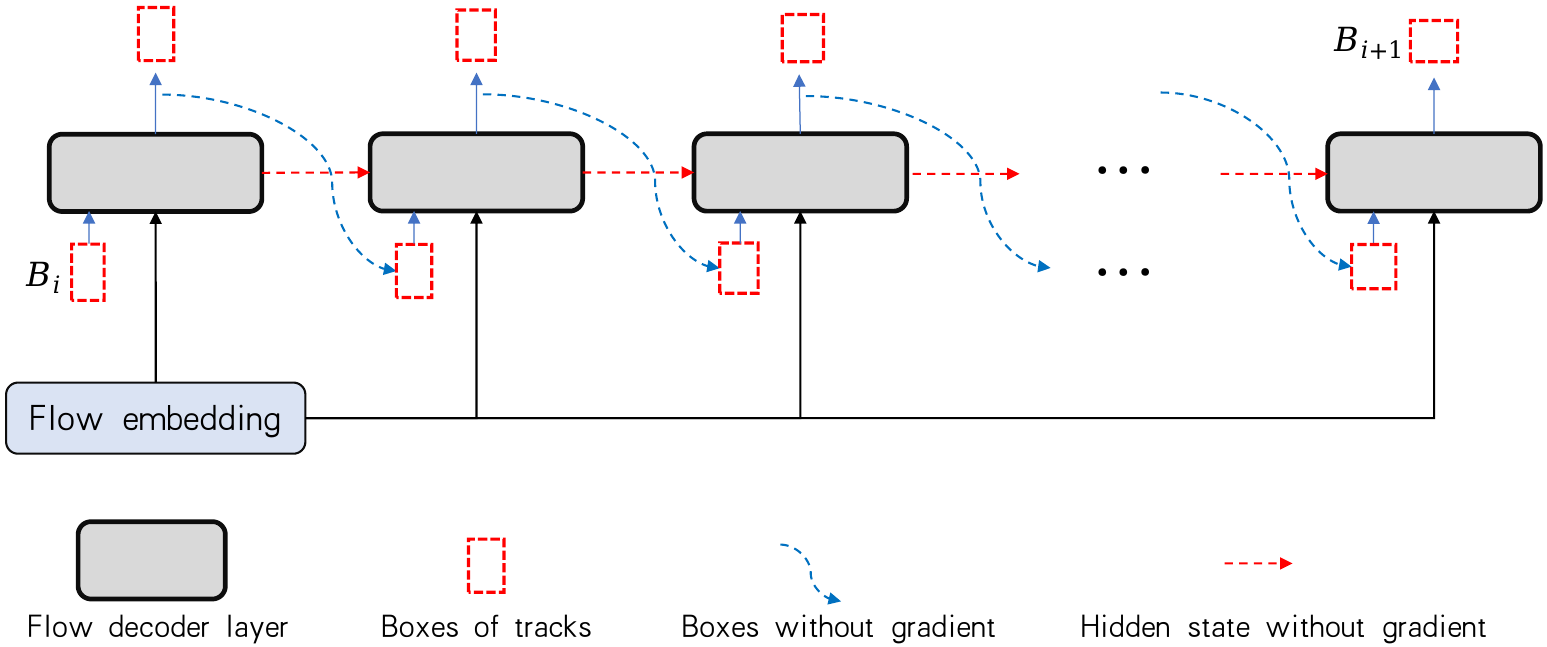}
\caption{
The overall structure of the flow decoder.
}
\label{dec}
\end{figure*}
\begin{figure}[!t]
\centering
\includegraphics[width=0.98\linewidth]{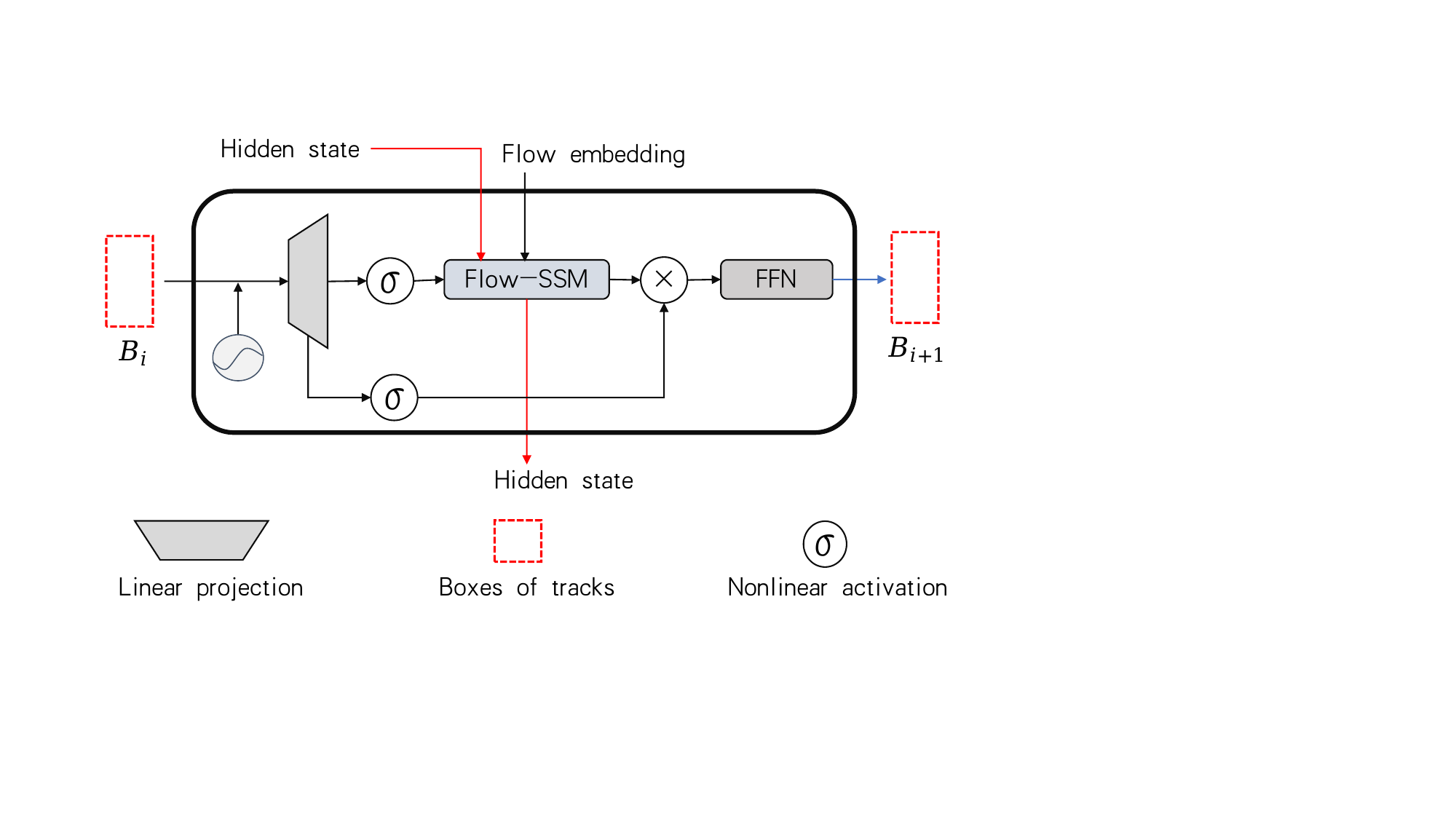}
\caption{
The structure of the each flow decoder layer.
}
\label{dec_layer}
\end{figure}
\subsection{Flow Decoder} 
To accurately regress the trajectory box from the previous frame to the current frame, we design a cascaded motion decoder, referred to as the flow decoder. The flow decoder is composed of $N$ identical decoder layers cascaded together, with each decoder layer containing a Flow-SSM module. The specific structure of the decoder layer is shown in \cref{dec_layer}. Given a trajectory box $B_i$, we apply cosine positional encoding to transform it into a high-dimensional trajectory position embedding $\mathbf{E}_i$. subsequently, we input the trajectory position embedding $\mathbf{E}_i$ into a linear layer and split it along the dimension into two position features, $\mathcal{E}_i$ and $\mathcal{R}_i$. The position feature $\mathcal{E}_i$ is fed into the Flow-SSM module, resulting in the trajectory prediction feature $\mathcal{E}_{i+1}$. The feature $\mathcal{R}_i$ serves as a residual component, passing through a nonlinear activation layer, and is then multiplied by the trajectory prediction feature $\mathcal{E}_{i+1}$, enriching the representation of $\mathcal{E}_{i+1}$. Finally, we input the $\mathcal{E}_{i+1}$	into a feed-forward network(FFN) \cite{vit} to obtain the trajectory prediction box $B_{i+1}$.

However, the trajectory prediction $B_{i+1}$ output by a single decoder layer is insufficient for achieving precise trajectory prediction. Therefore, we cascade $N$ identical decoder layers to construct the flow decoder, with the overall framework shown in \cref{dec}. The flow embeddings obtained from the Mamba encoder \cite{mamba} are applied to each decoder layer. The output trajectory box from the previous decoder layer serves as the input for the next decoder layer, allowing for precise refinement of the trajectory box position over time. Additionally, the hidden state acts as a messenger, transmitting the state of the trajectory box between the cascaded decoder layers, enabling the trajectory box to gradually regress according to the ground truth (GT) labels. The specific details of the regression process will be described in \cref{SL}.

\subsection{Step-by-Step Linear Training Strategy}\label{SL}
The flow decoder refines the trajectory box through a cascading process. In each flow decoder layer, the refinement of the trajectory box is guided by the flow features. Based on the intuition that the flow features have the same guiding effect across all decoder layers, we propose a step-by-step linear training strategy(S$^2$L). The core of S$^2$L is to linearly decompose the temporal autoregressive process of the trajectory box into $N$ simple regression steps. Specifically, given the trajectory box $B_i$ at time $i$, the flow decoder regresses $B_i$ to $B_{i+1}$. If there are $N$ flow decoder layers, the regression process from $B_i$ to $B_{i+1}$	can be linearly decomposed into $N$ sub-processes. It can be expressed as:
\begin{equation}
\label{decompose}
\begin{aligned}
\Delta_t = \frac{(i+1) - i}{N}, \\
\{B_{i+(k+1)\Delta_t} \leftarrow B_{i+k\Delta_t}\}^{N-1}_{k=0} =& I(B_{i+1} \leftarrow B_i),
\end{aligned}
\end{equation}
Where $\Delta_t$ is the time step, and $I$ is the linear interpolation function. By performing linear interpolation between $B_i$ and $B_{i+1}$, we can construct pseudo-labels $\{B_{i+k\Delta_t}\}^{N}_{k=1}$ for supervising each flow decoder layer. From the perspective of the entire regression process, by constructing pseudo-labels obtained through linear interpolation, we encourage the flow decoder to learn the regression process from $B_i$ to $B_{i+1}$ in a linear recursive manner, which can be expressed as:
\begin{equation}
\label{regression}
B_i \xrightarrow{\mathbf{f_1}(h, \mathcal{F})} B_{i+\Delta_t} \xrightarrow{\mathbf{f_2}(h, \mathcal{F})} \cdots \xrightarrow{\mathbf{f_N}(h, \mathcal{F})} B_{i+N\Delta_t},
\end{equation}
Where, $\mathbf{f}$ represents each flow decoder layer, $h$ is the hidden state, and $\mathcal{F}$ denotes the flow features. By using the step-by-step linear training strategy, the flow features guide the trajectory box through an equal amount of transformation in each decoder layer. This approach enables the flow decoder to handle more complex trajectory motions and improves the recall rate of lost trajectories.

\subsection{Training Loss of the TrackSSM}
We use ground truth as well as pseudo labels obtained through linear interpolation to supervise TrackSSM. We employ the smooth L1 loss and generalized intersection over union (GIoU) \cite{girshick14CVPR,giou} loss for training TrackSSM, specifically expressed as follows:
\begin{equation}
\label{loss}
\mathbf{\mathcal{L}}_{total} = \lambda_1 \mathbf{\mathcal{L}}_{smooth L1} + \lambda_2 \mathbf{\mathcal{L}}_{GIoU},
\end{equation}
Where $\lambda_1$ and $\lambda_2$ are the weight coefficients for the smooth L1 loss and GIoU loss, respectively. The specific formula for the smooth L1 loss is as follows:
\begin{equation}
\label{smooth_l1}
\mathbf{\mathcal{L}}_{smooth L1} = \left\{
\begin{aligned}
0.5(\mathbf{\hat{B}} - \mathbf{B})^2 & , &|\mathbf{\hat{B}} - \mathbf{B}| < 1, \\
|\mathbf{\hat{B}} - \mathbf{B}|-0.5& , & otherwise,
\end{aligned}
\right.
\end{equation}
The $\mathbf{\hat{B}}$ represents the trajectory box predicted by each decoder layer, and $\mathbf{B}$ represents the supervision labels. The matrix operations in \cref{smooth_l1} are element-wise.

\section{Experiments}\label{exp}
\subsection{Setting}
\subsubsection{Datasets}We evaluate the performance of TrackSSM in pedestrian, dancing, and sports scenarios, corresponding to the MOT17 \cite{mot16}, DanceTrack \cite{dancetrack}, and SportsMOT \cite{sportsmot} benchmarks, respectively. We merge the MOT17 and MOT20 \cite{mot20} training sets, referring to the combined set as MIX. For reporting results on the MOT17 test set, we train TrackSSM on MIX. For reporting results on the DanceTrack and SportsMOT test sets, we train TrackSSM separately on their respective training sets. For the ablation experiments, we train TrackSSM on the MIX, DanceTrack and SportsMOT training sets, respectively, and perform ablation testing on the MOT17 training set, DanceTrack validation set, and SportsMOT validation set.

\subsubsection{Metrics}We use the standard CLEAR metrics (MOTA, \etc) \cite{CLEAR-metrics}, HOTA \cite{hota}, AssA, DetA and IDF1 \cite{idf1} to comprehensively evaluate tracking performance. HOTA is an important metric for assessing the overall performance of detection and association. IDF1 is used to measure the precision and recall of trajectory associations. AssA and DetA specifically focus on measuring association accuracy and detection accuracy, respectively. Additionally, we use frames per second (FPS) to assess the efficiency of the tracker.

\subsubsection{Implementation details}
\textbf{Training.} During the training phase of TrackSSM, we train using trajectory segments rather than images, following the approach used in DiffMOT \cite{diffmot}. We select the position and motion information of historical frame trajectories with a time length of 5 as the input to TrackSSM. For the training setup, we set the default batch size to 2048 and use the Adam optimizer with a learning rate of 0.0001. The number of layers in the flow decoder is set to 6 by default. For the MIX, we train TrackSSM for a total of 160 epochs. For the DanceTrack \cite{dancetrack}, we train TrackSSM for 120 epochs. For the SportsMOT \cite{sportsmot}, we train TrackSSM for 340 epochs. Since some ground-truth boxes in the MOT17 \cite{mot16} exceed the image boundaries, we omit the bounding box normalization step when training on the MIX dataset. Additionally, when training TrackSSM on the MIX and DanceTrack, we use only the smooth L1 loss. However, when training TrackSSM on the SportsMOT, we use both the smooth L1 loss and GIoU \cite{giou} loss. It is taken because the object displacement distances between adjacent frames in the SportsMOT are greater, and using the GIoU loss helps the motion model converge more quickly.

\textbf{Inference.} During the inference phase, we set the default resolution of the input image to $800\times1440$ and use the publicly available YOLOX-x \cite{yolox} detector to infer detection results. To ensure a fair comparison with the baseline \cite{bytetrack}, we fix all hyper-parameter settings during tracking inference. The high-score detection threshold and low-score detection threshold are set to 0.6 and 0.1, respectively. For the non-maximum suppression (NMS) \cite{girshick14CVPR} post-processing, we fix the intersection over union (IoU) threshold at 0.7 and the confidence threshold at 0.01. The tracker use positional information for association, without the involvement of appearance features.

\textbf{Device.} We train TrackSSM with 2 GeForce RTX 3090 GPUs. During the inference phase, we perform tracking using a single GeForce RTX 3090 GPU.
 
\subsection{Evaluation of Different Benchmark}
We replace the Kalman filter \cite{kf} in ByteTrack \cite{bytetrack} with the TrackSSM motion model. For ease of reference, we temporarily refer to ByteTrack using TrackSSM as ByteSSM. We evaluate the ByteSSM on the MOT17 \cite{mot16}, DanceTrack \cite{dancetrack} and SportsMOT \cite{sportsmot} benchmarks to compare with other methods. The evaluation results are shown in \cref{tab:17}, \cref{tab:dance} and \cref{tab:sports}, respectively.\footnote{The best results are shown with bold in \cref{tab:17}, \cref{tab:dance} and \cref{tab:sports}.}.

\begin{table}[t]
\caption{The comparison of ByteSSM with other methods on the MOT17 test set. The $*$ indicates that this method is identical to ByteSSM in all settings except for the motion model.}
\label{tab:17}
\resizebox{1.0\linewidth}{!}{
\setlength{\tabcolsep}{1.0pt}
\begin{tabular}{ l | c c c c c c r}
\toprule
Tracker & HOTA$\uparrow$  & MOTA$\uparrow$ & IDF1$\uparrow$ &  AssA$\downarrow$ & DetA$\downarrow$ \\
\midrule
\textcolor{red}{\textbf{$kalman$ $filer$:}}\\
FairMOT\cite{fairmot} & 59.3 & 73.7 & 72.3 & 58.0 & 60.9\\
ByteTrack\cite{bytetrack} & 63.1 & 80.3 & 77.3 & 62.0 & 64.5 \\
ByteTrack$^*$\cite{bytetrack}&62.8 &78.7 &76.8 &62.1 &63.8\\
OC-SORT\cite{oc-sort} & 63.2 & 78.0 & 77.5 & 63.4 & 63.2 \\ 
SparseTrack\cite{sparsetrack}&65.1 &81.0 &80.1 &65.1 &65.3\\
\midrule
\textcolor{red}{\textbf{$learnable$ $motion$:}}\\
CenterTrack\cite{centertrack}&52.2 &67.8 &64.7 &51.0 &53.8\\
TraDes\cite{TraDeS}& 52.7 & 69.1 & 63.9 & 50.8 & 55.2 \\
QuasiDense\cite{qdtrack} & 53.9 & 68.7 & 66.3 & 52.7 & 55.6\\
TransTrack\cite{transtrack}& 54.1 & 75.2 & 63.5 &47.9 &61.6 \\
TransCenter\cite{transcenter}&54.5 &73.2 &62.2 &49.7 &60.1\\
TrackFormer\cite{trackformer}&57.3 &74.1 &68.0 &54.1 &60.9 \\
MeMOTR\cite{MeMOTR} &58.8 &72.8 &71.5 &58.4 &59.6 \\
GTR\cite{GTR}&59.1 &75.3 &71.5  &57.0 &61.6 \\
MOTR\cite{motrv2} &57.8 &73.4 &68.6 &55.7 & 60.3\\
STDFormer\cite{hu2023stdformer} & 60.9 & 78.4 & 73.1 &58.4 & -  \\
MambaTrack$^+$\cite{mambatrack+} &61.1 &78.1 &73.9 & - & - \\
ByteSSM &\textbf{61.4} &\textbf{78.5} &\textbf{74.1} & \textbf{59.6} & \textbf{63.6} \\
\bottomrule
\end{tabular}
}
\end{table}

\textbf{MOT17.} 
The MOT17 \cite{mot16} dataset contains frequent occlusions and slight camera movements, which pose a challenge to the ability of motion models to fit trajectories. By using TrackSSM as the motion model, ByteSSM achieves the highest level among many learnable motion methods. Compared to the baseline ByteTrack \cite{bytetrack}, ByteSSM's performance is slightly weaker. It is because the constant velocity motion prior in the kalman filter \cite{kf} naturally aligns well with the pedestrian movements in the MOT dataset \cite{mot15,mot16,mot20}.

\begin{table}[t]
\caption{The comparison of ByteSSM with other methods on the DanceTrack test set. The $*$ indicates that this method is identical to ByteSSM in all settings except for the motion model.}
\label{tab:dance}
\resizebox{1.0\linewidth}{!}{
\setlength{\tabcolsep}{1.0pt}
\begin{tabular}{ l | c c c c r}
\toprule
Tracker & HOTA$\uparrow$ & MOTA$\uparrow$ & IDF1$\uparrow$ & AssA$\uparrow$ & DetA$\uparrow$ \\
\midrule
\textcolor{red}{\textbf{$kalman$ $filter$:}}\\
FairMOT\cite{fairmot}  & 39.7 & 82.2 & 40.8 & 23.8 & 66.7 \\
ByteTrack$^*$\cite{bytetrack} & 46.8 & 89.6 & 51.8 & 30.9 & 71.3 \\
ByteTrack\cite{bytetrack} & 47.7 & 89.6 & 53.9 & 32.1 & 71.0 \\
BoT-SORT\cite{BoT-SORT} & 54.7 & 91.3 & 56.0 & 37.8 & 79.6 \\
OC-SORT\cite{oc-sort} & 55.1 & 92.0 & 54.6 & 38.3 & 80.3\\ 
SparseTrack\cite{sparsetrack} & 55.7 & 91.3 & 58.1 & 39.3 & 79.2 \\

\midrule
\textcolor{red}{\textbf{$learnable$ $motion$:}}\\
CenterTrack\cite{centertrack} & 41.8 &  86.8 & 35.7 & 22.6 & 78.1\\
TraDes\cite{TraDeS}& 43.3 & 86.2 & 41.2 & 25.4 & 74.5\\
TransTrack\cite{transtrack}& 45.5 & 88.4 & 45.2 & 27.5 & 75.9 \\
GTR\cite{GTR} & 48.0 & 84.7 & 50.3 & 31.9 & 72.5\\
QuasiDense\cite{qdtrack} & 54.2 & 87.7 & 50.4 & 36.8 & 80.1\\
MOTR\cite{motr} & 54.2  & 79.7 & 51.5 & 40.2& 73.5 \\
DiffMOT$^*$\cite{diffmot} &56.1 &92.2 &55.7 &38.5 &\textbf{81.9} \\
MambaTrack$^+$\cite{mambatrack+}&56.1 &90.3 &54.9 &39.0 &80.8 \\
ETTrack\cite{ettrack} &56.4 &92.2 &57.5 &39.1 &81.7 \\
MambaTrack\cite{mambatrack} &56.8 &90.1 &\textbf{57.8} &39.8 &80.1 \\
ByteSSM &\textbf{57.7} &\textbf{92.2} &57.5 &\textbf{41.0} &81.5 \\
\bottomrule
\end{tabular}
}
\end{table}

\textbf{DanceTrack.} 
DanceTrack \cite{dancetrack} is a more challenging benchmark for multi-object tracking with intense nonlinear motion, frequent occlusions, and similar appearances among objects. With the same detection and hyper-parameter settings, ByteSSM that adopts the TrackSSM motion model achieves the association gain of \textbf{+10.9} HOTA \cite{hota}, \textbf{+10.1} AssA and \textbf{+5.7} IDF1 \cite{idf1} compared to the baseline method. Among various trackers with learnable motion modules, ByteSSM achieves the best tracking performance. It demonstrates TrackSSM's exceptional ability to model nonlinear motion trajectories.

\begin{table}[t]
\caption{The comparison of ByteSSM with other methods on the SportsMOT test set. The $*$ indicates that this method is identical to ByteSSM in all settings except for the motion model.}
\label{tab:sports}
\resizebox{1.0\linewidth}{!}{
\setlength{\tabcolsep}{1.0pt}
\begin{tabular}{ l | c c c c r}
\toprule
Tracker & HOTA$\uparrow$ & MOTA$\uparrow$ & IDF1$\uparrow$ & AssA$\uparrow$ & DetA$\uparrow$ \\
\midrule
\textcolor{red}{\textbf{$kalman$ $filter$:}}\\
FairMOT\cite{fairmot} &49.3 &86.4 &53.5 &34.7 &70.2  \\
ByteTrack\cite{bytetrack}&64.1 &95.9 &71.4 &52.3 &78.5 \\
ByteTrack$^*$\cite{bytetrack} &63.4 &95.7 &70.3 &51.3 &78.4 \\
OC-SORT\cite{oc-sort}&73.7 &96.5 &74.0 &61.5 &88.5 \\ 

\midrule
\textcolor{red}{\textbf{$learnable$ $motion$:}}\\
GTR\cite{GTR}&54.5 &67.9 &55.8 &45.9 &64.8   \\
QuasiDense\cite{qdtrack}&60.4 &90.1 &62.3 &47.2 &77.5  \\
CenterTrack\cite{centertrack} &62.7 &90.8 &60.0 &48.0 &82.1 \\
TransTrack\cite{transtrack}&68.9 &92.6 &71.5 &57.5 &82.7 \\
MeMOTR\cite{MeMOTR}&70.0  &91.5  &71.4  &59.1  &83.1   \\
MambaTrack$^+$\cite{mambatrack+}&71.3  &94.9 &71.1  &58.6  &86.7   \\
MambaTrack\cite{mambatrack} &72.6 &95.3 &72.8 &60.3 &87.6  \\
DiffMOT$^*$\cite{diffmot}&74.0 &96.8 &73.7 &61.7 &\textbf{88.9} \\
MotionTrack\cite{motionTrack} &74.0 &96.6  &74.0  &61.7 &88.8  \\
MixSort-OC\cite{sportsmot} &74.1 &96.5 &74.4 &62.0 &88.5   \\
ETTrack\cite{ettrack} &74.3 &96.8 &74.5 &62.1  &88.8   \\
ByteSSM &\textbf{74.4} &\textbf{96.8} &\textbf{74.5} &\textbf{62.4}  &88.8   \\
\bottomrule
\end{tabular}
}
\end{table}

\textbf{SportsMOT.} 
SportsMOT \cite{sportsmot} is a large-scale multi-object tracking benchmark designed for sports scenarios. It includes three types of sports scenes: soccer, basketball, and volleyball. Compared to the MOT challenge benchmark \cite{mot15,mot16,mot20}, SportsMOT presents diverse motion patterns and complex nonlinear movements, posing a significant challenge to the tracker's ability to model complex trajectory motions. With the same detection and hyper-parameter settings, ByteSSM achieves the association gain of \textbf{+11.0} HOTA, \textbf{+11.1} AssA and \textbf{+4.2} IDF1 compared to the baseline. Among various tracking methods utilizing learnable motion modules, ByteSSM continues to maintain the best tracking performance. It strongly demonstrates the potential of TrackSSM as a universal motion predictor.

\subsection{Ablations}
\subsubsection{The impact of different motion models on tracking performance}
By keeping the detection and hyper-parameter settings fixed, we replace the Kalman filter \cite{kf} motion module in ByteTrack \cite{bytetrack} with TrackSSM and DiffMOT \cite{diffmot} in succession to observe the performance differences between different motion models. We conduct inference testing using the aforementioned motion modules on the MOT17 \cite{mot16} training set, DanceTrack \cite{dancetrack} validation set and SportsMOT \cite{sportsmot} validation set. The results are shown in \cref{ablation:motion-compare}.

In pedestrian scenarios, the kalman filter with a constant velocity motion prior achieves the best tracking results. It is because most pedestrian objects move at approximately constant speeds. However, in dancing and sports scenarios, the motion of the objects becomes nonlinear and the targets undergo significant deformations. In this cases, the constant velocity assumption no longer provides a reliable positional prior, leading to weaker tracking performance. Compared to the Kalman filter, TrackSSM achieves comparable performance in pedestrian scenarios and significantly outperforms the kalman filter motion model in dancing and sports scenarios. It indicates that learnable motion models can better fit the nonlinear and non-rigid motions of trajectories, thereby improving association accuracy during tracking. Moreover, when compared to DiffMOT, which is also a learnable motion model, TrackSSM consistently outperforms DiffMOT across all three scenarios. It proves the ability of TrackSSM that handle nonlinear motion trajectories and its broad applicability across different scenarios.

\begin{table*}[!ht]
\caption{
Comparison of the tracking performance with different motion modules.
}
\label{ablation:motion-compare}
\setlength{\tabcolsep}{7pt}
\renewcommand\arraystretch{1.2}
\centering
\begin{tabular}{ c c |c c c |c c c |c c c}
\toprule
\multirow{2.5}{*}{Motion models} & &\multicolumn{3}{c}{MOT17}  & \multicolumn{3}{c}{DanceTrack} & \multicolumn{3}{c}{SportsMOT} \\
\cmidrule(r){3-5} \cmidrule(lr){6-8} \cmidrule(lr){9-11}
 & & HOTA$\uparrow$ & AssA$\uparrow$  & IDF1$\uparrow$ & HOTA$\uparrow$ & AssA$\uparrow$ & IDF1$\uparrow$ & HOTA$\uparrow$ & AssA$\uparrow$ & IDF1$\uparrow$ \\
\midrule
kalman filter&&\textbf{75.0} &\textbf{72.4} &\textbf{83.3} &47.1 &31.4 &51.0 &67.8 &57.2 &76.0  \\
DiffMOT      &&73.9 &69.6 &81.3 &53.7 &36.7 &53.3 &80.9 &70.1 &80.3  \\
TrackSSM     &&74.9 &71.4 &81.7 &\textbf{53.8} &\textbf{36.8} &\textbf{53.7} &\textbf{81.2} &\textbf{70.7} &\textbf{80.8}  \\
 
\bottomrule
\end{tabular}
\end{table*}

\subsubsection{The impact of trajectory segment with different lengths on tracking performance}
We explore the impact of the length of trajectory segments input to TrackSSM on tracking performance. Specifically, during the training phase, we train trajectory segments with historical time lengths of 3, 5, 10, 20 and 40, separately, while keeping the training settings consistent. During the inference phase, we use trajectory information with historical time lengths of 3, 5, 10, 20 and 40 for motion prediction, separately, again maintaining consistent detection and hyper-parameter settings. The experimental results are shown in \cref{ablation:length-compare}.

When the historical time length of trajectory segments used in both the training and inference phases is set to 3, ByteSSM achieves the best tracking performance. It suggests that trajectory information closer to the current time is more relevant to future trajectory predictions. For the DanceTrack \cite{dancetrack} dataset, the tracking performance of ByteSSM continues to improve as the historical time length of the input trajectory segments increases. We speculate that the movement and actions of targets in dancing scenes often exhibit a certain degree of periodicity. As the historical trajectory information increases, TrackSSM may learn the complete motion cycle of the tracked target, which facilitates accurate prediction of the future position of dancers.

\begin{table}[!t]
\caption{
\lzl{
Comparison of the tracking performance with different lengths of trajectory segments.
}
}
\label{ablation:length-compare}
\renewcommand\arraystretch{1.1}
\centering
\setlength{\tabcolsep}{4pt}
\begin{tabular}{c c |c c c |c c c}
\toprule
\multirow{2.55}{*}{Lengths}& & \multicolumn{3}{c}{DanceTrack}  & \multicolumn{3}{c}{SportsMOT} \\
\cmidrule(r){3-5} \cmidrule(lr){6-8}
 & & HOTA$\uparrow$ & AssA$\uparrow$  & IDF1$\uparrow$ & HOTA$\uparrow$ & AssA$\uparrow$ & IDF1$\uparrow$ \\
\midrule
3           & &\textbf{53.9} &\textbf{36.9} &53.0   &\textbf{81.5}   &\textbf{71.2}   &\textbf{81.2}  \\
5           & &53.8 &36.8 &\textbf{53.7}   &81.2   &70.7   &80.8  \\
10          & &52.5 &34.9 &50.0   &81.1   &70.6   &80.6  \\
20          & &52.9 &35.6 &52.3   &80.9   &70.2   &80.1  \\
40          & &53.9 &36.8 &52.8   &80.6   &69.6   &79.6  \\
\bottomrule
\end{tabular}
\end{table}

\subsubsection{The impact of different numbers of decoder layers on tracking performance
}
We further explore the impact of the number of decoder layers on tracking performance. We fix all training settings and train TrackSSM with 1, 2, 3, 6 and 12 decoder layers, respectively. During the inference phase, we fix all detection and hyper-parameter settings and use TrackSSM with 1, 2, 3, 6, and 12 decoder layers for motion prediction, separately. The experimental results are shown in \cref{ablation:layers-compare}.

For the flow decoder with a single layer, the training process involves directly predicting the trajectory's future position without the need for a step-by-step linear prediction process. Despite this, the flow decoder with one layer still achieve decent tracking performance, indicating that even a single flow decoder layer can perform trajectory prediction with reasonable accuracy. When the number of decoder layers is set to 6, ByteSSM achieves the best tracking performance. Therefore, we select the flow decoder with 6 layers as the default configuration for TrackSSM.

\begin{table}[!t]
\caption{
Comparison of the tracking performance with different numbers of decoder layers.
}
\label{ablation:layers-compare}
\renewcommand\arraystretch{1.1}
\centering
\setlength{\tabcolsep}{4pt}
\begin{tabular}{c c |c c c |c c c}
\toprule
\multirow{2.55}{*}{Layers}& & \multicolumn{3}{c}{DanceTrack}  & \multicolumn{3}{c}{SportsMOT} \\
\cmidrule(r){3-5} \cmidrule(lr){6-8}
 & & HOTA$\uparrow$ & AssA$\uparrow$  & IDF1$\uparrow$ & HOTA$\uparrow$ & AssA$\uparrow$ & IDF1$\uparrow$ \\
\midrule
1           & &53.4  &36.3  &52.8  &80.3  &69.2 &79.4  \\
2           & &52.8  &35.3  &50.8  &80.0  &68.8  &79.0    \\
3           & &52.4  &34.8  &51.2  &79.6  &68.0  &78.6    \\
6           & &\textbf{53.8}  &\textbf{36.8}  &\textbf{53.7}  &\textbf{81.2}  &\textbf{70.7}  &\textbf{80.8}    \\
12          & &52.8  &35.2  &51.0  &80.6  &69.7  &80.1      \\
\bottomrule
\end{tabular}
\end{table}

\subsubsection{The impact of the step-by-step linear training strategy on tracking performance
}
To observe the impact of the step-by-step linear training strategy (S$^2$L) on TrackSSM's motion modeling, we conduct an ablation analysis for S$^2$L on the DanceTrack \cite{dancetrack} and SportsMOT \cite{sportsmot} validation datasets, separately. We perform training plans with and without the S$^2$L strategy, keeping all other configurations unchanged. The experimental results are shown in \cref{ablation:s2l-compare}.

Clearly, after applying the S$^2$L strategy, the association performance of trackers improved across validation sets in both scenarios. On the DanceTrack validation set, TrackSSM trained with the S$^2$L strategy yields performance gains of +2.4 HOTA \cite{hota}, +3.3 AssA, and +4.5 IDF1 \cite{idf1}. It indicates that the S$^2$L strategy helps TrackSSM more accurately regress nonlinear motion trajectory boxes and recalls lost trajectories (as evidenced by the significant improvement in the IDF1 metric). On the SportsMOT validation set, TrackSSM trained with the S$^2$L strategy still provides the tracker with performance gains of +0.3 HOTA, +0.3 AssA, and +0.4 IDF1. It demonstrates the general applicability of the S$^2$L strategy across different scenarios. Notably, the S$^2$L strategy brings more substantial gains to the tracker on the DanceTrack dataset. We speculate that it is due to two factors: 1) The DanceTrack dataset contains more non-rigid motions. 2) The S$^2$L strategy focuses on handling non-rigid deformation motion of trajectory boxes, which is a type of nonlinear motion. By linearly decomposing non-rigid deformations into several simple, equal transformation steps, the S$^2$L strategy enables the flow decoder to more easily learn the non-rigid motion of trajectory boxes.

\begin{table}[!t]
\caption{
Comparison of the tracking performance with or without the step-by-step linear training strategy.
}
\label{ablation:s2l-compare}
\renewcommand\arraystretch{1.1}
\centering
\setlength{\tabcolsep}{3.5pt}
\begin{tabular}{c c |c c c |c c c}
\toprule
\multirow{2.55}{*}{w/ S$^2$L}& & \multicolumn{3}{c}{DanceTrack}  & \multicolumn{3}{c}{SportsMOT} \\
\cmidrule(r){3-5} \cmidrule(lr){6-8}
 & & HOTA$\uparrow$ & AssA$\uparrow$  & IDF1$\uparrow$ & HOTA$\uparrow$ & AssA$\uparrow$ & IDF1$\uparrow$ \\
\midrule
             &&51.4  &33.5 &49.2  &80.9  &70.4  &80.4   \\
\checkmark   &&\textbf{53.8}  &\textbf{36.8} &\textbf{53.7} &\textbf{81.2}  &\textbf{70.7}  &\textbf{80.8}   \\
\bottomrule
\end{tabular}
\end{table}

\subsubsection{The impact of different detectors on tracking performance}
We use publicly available different versions of the YOLOX \cite{yolox} detector for inference on the DanceTrack \cite{dancetrack} test set and employ TrackSSM with fixed weights for motion prediction. The experimental results are shown in \cref{ablation:eff-compare}. The parameter count of TrackSSM is indicated in blue text.

When a more lightweight detector is used, the accuracy of the tracker decreases, which aligns with the common pattern observed in tracking-by-detection (TBD) paradigms \cite{Bewley2016_sort}. It should be worth that the TrackSSM model has only 5.1M parameters, indicating its potential for deployment on edge devices. Additionally, when using the YOLOX-l detector, ByteSSM achieves a great trade-off between efficiency and tracking accuracy, with a running speed of 27.5 FPS, enabling real-time multi-object tracking.

\begin{table}[!t]
\caption{
Comparison of the tracking performance of TrackSSM adopt different detectors.
}
\label{ablation:eff-compare}
\renewcommand\arraystretch{1.2}
\centering
\setlength{\tabcolsep}{5pt}
\begin{tabular}{c c | c c c}
\toprule
Methods &    &FPS &\#param. & HOTA  \\
\midrule
TrackSSM+YOLOX-x&&20.3  &104M(\textcolor{blue}{5.1M})& 57.7  \\
TrackSSM+YOLOX-l&&27.5  &59.2M(\textcolor{blue}{5.1M})&57.3    \\
TrackSSM+YOLOX-m&&29.8  &30.4M(\textcolor{blue}{5.1M})&52.8   \\
TrackSSM+YOLOX-s&&31.9  &14.1M(\textcolor{blue}{5.1M})&48.5    \\
\bottomrule
\end{tabular}
\end{table}

\section{Conclusion}\label{conclusion}
We propose a simple and efficient motion model with an encoder-decoder structure, named TrackSSM. It uses a naive Mamba module to build the encoder, which converts the position and motion information of historical trajectories into flow features. In the decoding phase, to enhance the ability of fitting nonlinear motion, we introduce Flow-SSM. It uses flow features as a guide, facilitating precise temporal autoregression of the trajectory boxes. To further improve the accuracy of trajectory prediction, we carefully design the flow decoder, which is composed of several identical decoder layers cascaded together to refine the trajectory boxes step by step. Additionally, we propose a step-by-step linear training strategy (S$^2$L), which linearly decomposes the regression process of the trajectory boxes into several simple transformation steps. This strategy enhances TrackSSM's ability to model lost trajectories and complex motion trajectories. Compared to the popular kalman filter \cite{kf} motion model, TrackSSM adapts to object motion in various scenarios and provides precise trajectory predictions for trackers. When compared to motion models using attention mechanisms \cite{transtrack,transcenter,trackformer,transmot,motr,motrv2}, TrackSSM achieves significant motion prediction capabilities with much lower computational overhead, demonstrating its efficiency and robustness. In the future, we will continue to explore the potential of SSM-like tracking models in both the spatial-temporal dimensions, not just the temporal dimension. We also hope that this work will inspire the design of SSM-based decoder structures and anticipate the development of more elegant methods.
\bibliographystyle{IEEEtran}
\bibliography{6_ref}

\newpage

\end{document}